\pdfoutput=1

\documentclass[11pt]{article}

\usepackage[]{emnlp2021}

\usepackage{times}
\usepackage{latexsym}

\usepackage[T1]{fontenc}

\usepackage[utf8]{inputenc}

\usepackage{microtype}

\newcommand{\ignore}[1]{}
\newcommand{\nop}[1]{}

\newcommand{\eg}{\textit{e.g.,~}}
\newcommand{\ie}{\textit{i.e.,~}} 

\usepackage{array}

 \usepackage{graphics}
 \usepackage[export]{adjustbox}
 \usepackage{enumitem}
\usepackage[tight,footnotesize]{subfigure}
 
\usepackage{amsmath}

\usepackage{pgfplots} 
\usepackage{tikz}

\usepackage{multirow, multicol}
 
\usepackage{xcolor,colortbl}

\colorlet{posColor}{green!60!yellow!70!black}
\colorlet{negColor}{magenta}

\usepackage{amsfonts}

\title{Identifying Causal Influences on Publication Trends and Behavior: \\A Case Study of the Computational Linguistics Community}

\author{
        Maria Glenski  \and Svitlana Volkova  \\
        National Security Directorate \\ Pacific Northwest National Laboratory \\ 
        \texttt{\{Maria.Glenski@pnnl.gov, Svitlana.Volkova@pnnl.gov\}}
        }

\begin{document}
\maketitle
\begin{abstract}

Drawing causal conclusions from observational real-world data is a very much desired but  challenging task. In this paper we present mixed-method analyses to investigate causal influences of publication trends and behavior on the adoption, persistence, and retirement of certain research foci -- methodologies, materials, and tasks that are of interest to the computational linguistics (CL) community. Our key findings highlight evidence of the transition to rapidly emerging methodologies in the research community (\eg adoption of bidirectional LSTMs influencing the retirement of LSTMs), the persistent engagement with trending tasks and techniques (\eg  deep learning, embeddings, generative, and language models), the effect of scientist location from outside the US, \eg China on propensity of researching languages beyond English, and the potential impact of funding for large-scale research programs. We anticipate this work to provide useful insights about publication trends and behavior and raise the awareness about the potential for causal inference in the computational linguistics and a broader scientific community.  
 
\end{abstract}

 \begin{figure}[t!]
     \centering
     \includegraphics[width=0.4\textwidth]{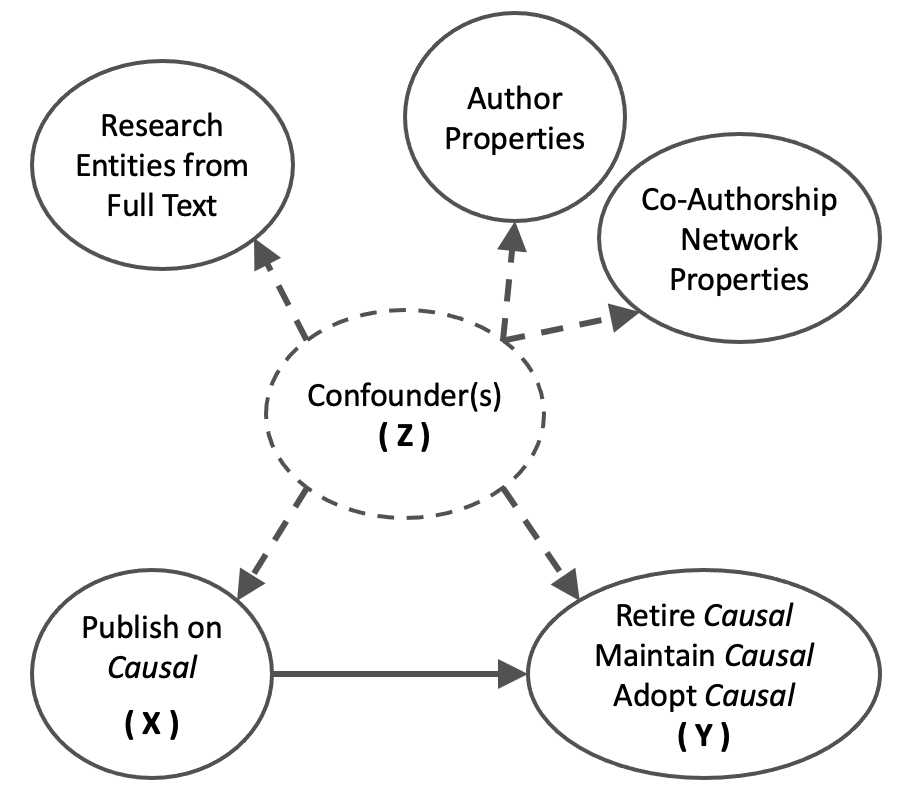}
     \caption{{Causal diagram for scientific publications, where full text, author, and co-citation network properties encode causal confounders (e.g., author influence, research tasks and methodology) used as covariates. Our analyses rely on the assumption that latent confounders can be measured and adjusted for based on proxies in our covariates.}}
     \label{fig:causal_diagram}
 \end{figure}

Causal understanding is essential for informed decision making~\cite{pearl2018book,Pearl2019Seven,varian2016causal} to go beyond correlations and overcome the predictability limit of real-world partially observed systems including complex systems of human social behavior~\cite{abeliuk2020predictability,hofman2017prediction}. 

Unlike earlier work that focused on analysing publication trends, diversity and innovation in science relying on descriptive exploratory analysis primarily driven by correlations~\cite{fortunato2018science,hofstra2020diversity,ramage2020mapping}, this work aims to provide empirical evidence of causal mechanisms driving publication trends and behavior in the computational linguistics community. Our key contributions are two-fold. First, we experiment and evaluate the potential of using complementary causal inference approaches, specifically causal structure learning models and several treatment effect estimation techniques to measure causal influences in high-dimensional observational data. Second, we analyze the temporal dynamics of causal influences of publication trends on the adoption, persistence and retirement of certain research foci in the CL community.  

For our analyses we leverage public ACL anthology data and encode publication and scientist characteristics, as well as collaboration behavior as confounders, to measure the effect of previous research foci on the adoption, maintenance and retirement of future research foci focusing on most recent six years between 2014 and 2019.  
Figure~\ref{fig:causal_diagram} presents a causal diagram which illustrates our core assumption: that latent confounders (\eg author reputation, productivity, and collaborations, the strength and length of authors' research careers, novelty of research, papers' contributions to the field, etc.) can be measured and adjusted for based on proxies in our covariates (\eg research entities extracted from publications, author properties including time since first paper, volume of papers, centrality within co-authorship networks etc.).
 
\section{Related Work}

There are two complementary causal inference frameworks -- structural causal models~\cite{pearl2009causal} and treatment effect estimation approaches~\cite{rosenbaum1983central}. 
Existing approaches to learn the causal structure (aka causal discovery) broadly fall into two categories: constraint-based~\citep{spirtes2000causation,yu2016review} and score-based~\citep{chickering2002optimal}. 

Recently, there have been an increased interest in causal inference on observational data~\cite{guo2020survey}, including text data, in the computational linguistics and computational social science communities~\cite{lazer2009social}. For example, recent work by~\cite{roberts2020adjusting} estimated the effect of perceived author gender on the number of citations of the scientific articles and~\cite{veitch2020adapting} measured the effect that presence of a theorem in a paper had on the rate of the paper's acceptance. 

Additional examples in the computational social science domain include: measuring the effect of alcohol mentions on Twitter on college success~\cite{kiciman2018using}; estimating the effect of the ``positivity'' of the product reviews and recommendations on sales on Amazon~\cite{pryzant2020causal,sharma2015estimating}; understanding factors effecting user performance on StackExhange, Khan Academy, and Duolingo~\cite{alipourfard2018using,fennell2019predicting}; 
estimating the effect of censorship on subsequent censorship and posting rate on Weibo~\cite{roberts2020adjusting} and word use in the mental health community on users' transition to post in the suicide community on Reddit~\cite{de2016discovering,de2017language}; or the effect of exercise on shifts of topical interest on Twitter~\cite{falavarjani2017estimating}. Moreover,~\citet{keith2020text} presented the overview of causal approaches for computational social science problems focusing on the use of text to remove confounders from causal estimates, which was also used by~\cite{weld2020adjusting}. Earlier work utilized matching methods to learn causal association  between word features and class labels in document classification~\cite{paul2017feature,wood2018challenges}, and use text as treatments~\cite{fong2016discovery,dorie2019automated} or covariates e.g., causal embeddings~\cite{veitch2020adapting,scholkopf2021toward}. 

Several studies have leveraged the ACL Anthology dataset to analyze diversity in computer science research~\cite{vogel2012he}, and perform exploratory data analysis such as knowledge extraction and mining~\cite{singh2018cl,radev2012rediscovering,gabor2016semantic}. However, unlike any other work, our approach focuses on leveraging complementary methods for causal inference -- structural causal models and treatment effect estimation to discover and measure the effect of scientists' research focus on their productivity and publication behavior, specifically the emergence, retirement and persistence of computational linguistics methodologies, approaches and topics.

\section{Data Preprocessing}
 
Our causal analysis relies on the publication records from the Association of Computational Linguistic (ACL) research community from 1986 through 2020. We collect the ACL Anthology dataset\footnote{\url{https://aclanthology.org/}}~\cite{gildea2018acl} with the bibtex provided with the accompanying abstracts. Excluding all records that do not contain authors (\eg bibtex entries for the workshop proceedings), we convert the bibtex representation into a data representation  
where each row represents each paper-author combination (\ie for a paper (paperX) with three authors, there are three representative rows: paperX-author1, paperX-author2, and paperX-author3). 

Then, we extract features that encode {\bf paper} properties: the year it was published, whether the paper was published in a conference or journal, the number of authors, the number of pages, and word count in paper. We also compute Gunning fog index~\cite{gunning1952technique} -- influenced by the number of words, sentences, and complex words. 
 
We then annotate each row with properties related to the {\bf author} during the year the paper was published. As a proxy of the length of the author's research career in the computational linguistics community, we calculate the number of years since the author's first publication in the anthology. Each author's location is represented as the location (country) of the institution the author is associated with in the metadata or full text.  
To measure productivity at varying granularities, we calculate the number of one's papers published in total, in the last year, and in the last five years.

We then construct a dynamic network representation of the anthology using author-to-paper relationships for each calendar year, as encoded in the metadata. After projecting those relationships into the dynamic co-authorship network that reflects author to co-author connections by year, we calculate centrality and page rank network statistics over time to measure the influence of the author. These {\bf collaboration behavior} features complement previously described author properties.
We also added three features to encode the diversity in co-authorship. First, the number of all co-authors who published the papers with the author. Second, the average number of papers co-authored per co-author, which is computed as the total number of papers co-authored per co-author divided by the number of co-authors. The last is a likelihood that a co-author is an author on a paper, which is the second feature divided by the total number of the author's papers. This enables us to measure the diversity, or lack thereof, of collaborative relationships of each author, and encodes how collaboration behavior evolves over time.

\subsection{Encoding Research Focus}
\label{sec:research_entities}
After extracting the full text of each paper from the PDF using GROBID~\cite{GROBID}, we use the SpERT model trained to extract key research entities from scientific publications. The SpERT model~\cite{luan2018multi} was trained to extract scientific entities of different types such as tasks, methods, and materials and the relationships between them such as ``Feature-Of" and ``Used-for", using the SciERC dataset\footnote{\url{http://nlp.cs.washington.edu/sciIE/}}. After applying the model to the ACL data, we consolidate noisy references of research entities into representative clusters manually, resulting in 50 entities that encode research tasks, methods, and materials\footnote{Research entities trending in the CL community used for our causal analyses: ``artificial intelligence'',
``adversarial'',
``annotation'',
``arabic'',
``attention'',
``baselines'',
``bidirectional lstm'',
``causal'',
``chinese'',
``classification'',
``coreference'',
``crowdsourcing'',
``deep learning'',
``dialog'',
``embeddings'',
``ethics'',
``explanation'',
``fairness'',
``french'',
``generative'',
``german'',
``grammars'',
``graph models'',
``heuristics'',
``interpretability'',
``language models'',
``lstm'',
``machine learning'',
``monolingual'',
``multilingual'',
``multiple languages'',
``NER'',
``node2vec'',
``non-English language'',
``pos/dependency/parsing'',
``QA'',
``reinforcement learning'',
``robustness'',
``russian'',
``sentiment'',
``statistical/probabilistic models'',
``summarization'',
``topic model'',
``transfer learning'',
``transformers'',
``translation'',
``transparency'',
``unsupervised methods'',
``word2vec'',
``benchmark''.}. 

\begin{figure}
    \centering
    \small
    \begin{tikzpicture}
 
    \begin{axis} [
    height=1.25in,
    width=3in, 
    xtick={1960,1970,1980,1990,2000,2010,2020},
    xticklabels={1960,1970,1980,1990,2000,2010,2020},
    xmin=1960, 
    xmax=2020,
    ylabel=\% Papers,
    ylabel near ticks 
    ]
    \addplot[no marks] coordinates {	
    (1965, 21.428571428571427)
	(1967, 50.0)
	(1969, 11.11111111111111)
	(1973, 0.0)
	(1974, 50.0)
	(1975, 16.666666666666664)
	(1976, 57.14285714285714)
	(1977, 9.090909090909092)
	(1978, 13.157894736842104)
	(1979, 37.5)
	(1980, 28.688524590163933)
	(1981, 31.428571428571427)
	(1982, 28.187919463087248)
	(1983, 31.292517006802722)
	(1984, 24.705882352941178)
	(1985, 25.146198830409354)
	(1986, 24.269005847953213)
	(1987, 29.11392405063291)
	(1988, 26.180257510729614)
	(1989, 34.62732919254658)
	(1990, 30.668414154652684)
	(1991, 30.25)
	(1992, 27.056277056277057)
	(1993, 32.241153342070774)
	(1994, 31.852551984877124)
	(1995, 35.80705009276438)
	(1996, 27.653061224489793)
	(1997, 33.56890459363957)
	(1998, 31.5018315018315)
	(1999, 34.35374149659864)
	(2000, 39.712606139777925)
	(2001, 36.22291021671827)
	(2002, 37.65541740674956)
	(2003, 35.54347826086956)
	(2004, 36.07907742998353)
	(2005, 41.518275538894095)
	(2006, 38.26611622737377)
	(2007, 39.38782374705684)
	(2008, 40.30816640986132)
	(2009, 40.278729723554946)
	(2010, 38.31394162073893)
	(2011, 37.28967712596635)
	(2012, 37.6173285198556)
	(2013, 37.87696019300362)
	(2014, 37.278106508875744)
	(2015, 40.542159652538565)
	(2016, 43.40630564575259)
	(2017, 46.94185753838409)
	(2018, 46.022632717590284)
	(2019, 47.5609756097561) 
    };

    \addplot[no marks,dashed] coordinates {
    (2014, 0)
	(2014, 58)
	};
	
    \end{axis}

    \end{tikzpicture}
 
    \caption{Relative coverage of consolidated research entity representations in the ACL data. Percentage of papers with at least one entity associated by publication year. Dashed line indicates the start of our causal analysis period (2014).}
    \label{fig:keyword_coverage}
\end{figure}
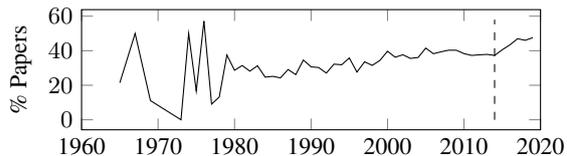

These consolidated entities are representative of the top 300 entities extracted from all ACL anthology publications for which we were able to extract the full text (121,134 out of 127,041 which is 95.3\% of all records for which there was an ACL anthology bibtex entry), after removing trivial or general terms such as ``system'', `approach'', ``it'', ``task'', and ``method''. We present the coverage across papers (\% of papers with at least one associated entity) over time in Figure~\ref{fig:keyword_coverage}, illustrating the coverage approximates the overall coverage (around 41\%) for the bulk of the dataset (1980-2019), with coverage trending upwards over time.

\section{Methodology}
We investigate the causal relationships between characteristics of individual publications 
for  
each researcher who authored at least two publications hosted in the ACL Anthology. We identify causal influences of publication outcomes related to the adoption, retirement, and maintenance of computational linguistic methodologies, approaches, or topics, as well as the productivity of authors using both causal discovery and pairwise causal effect estimation. 

\subsection{Treatments and Outcomes}

We consider the inclusion of our consolidated research entities to be treatments in our analyses --- \eg does the inclusion of \textit{biLSTM} architectures in the publication have a causal relationship with future research outcomes? --- and the basis of several research outcomes related to the \textit{adoption}, \textit{retirement}, and \textit{maintenance} of CL methodologies, tasks and approaches.

That is, the association of the identified research entities with authors' publications allows us to identify when authors {adopt} new emerging technologies (\eg the first use of \textit{transformers}), {retire} previously used methods or research applications (\eg if authors stop publishing on \textit{LSTM} architectures after \textit{biLSTM} architectures are introduced), continue to use -- or {maintain} publications in -- methods (\eg when authors continue to publish on \textit{NER}). We associate these behaviors as future outcomes for each author's publications in previous years. 

For each year in which an author published in an ACL venue, we calculate adoption and retirement outcomes for each consolidated research element the following year, maintenance outcomes for each research element considering the following two years.
Alongside these fine-grained research outcomes, we also examine coarse-grained, or general outcomes for authors:
\begin{itemize}[noitemsep,nolistsep]
    \item overall pauses in publishing within ACL venues (no publications in any ACL community for two years),
    \item persistent publication records (continuing to publish in consecutive years), 
    \item publication volume increases (the increase or decrease in number of publications in ACL venues relative to the previous year).
\end{itemize} 
 
In our analyses, we focus on recent six years (2014-2019) for which we have complete treatment and outcome annotations and consider each year independently. We leverage two types of publication record granularities -- publication records and yearly research portfolios -- to analyze the temporal dynamics of the causal system underpinning CL publication venues at multiple resolutions. Note, we present a detailed description of the treatments, covariates and outcomes we used  
in Appendix~\ref{sec:appendix}.

\subsection{Causal Structure Learning} 
Structural causal models are a way of describing relevant features of the world and how they interact with each other. Essentially, causal models represent the mechanisms by which data is generated. The causal model formally consists of two sets of variables $U$ (exogenous variables that are external to the model) and $V$ (endogenous variables that are descendants of exogenous variables), and a set of functions $f$ that assign each variable in $V$ a value based on the values of the other variables in the model. To expand this definition: a variable $X$ is a direct cause of a variable $Y$ if $X$ appears in the function that assigns $Y$ value. Graphical models or Directed Acyclic Graphs (DAGs) have been widely used as causal model representations. 

The causal effect rule is defined as: given a causal graph $G$ in which a set of variables $PA(X)$ are designated as a parents of $X$, the causal effect of $X$ on $Y$ is given by:
\begin{equation}
  \begin{array}{l}
P(Y=y | do (X=x)) =\\
\sum_{z} P (Y = y | X = x, PA = z) P(PA = z),
\end{array}
\end{equation}

where $z$ ranges over all the combinations of values that the variables in PA can take. 

The first approach for our causal analysis aims to examine the causal relationships that are identified using an ensemble of causal discovery algorithms~\cite{saldanhaevaluation}. Our ensemble considers the relationships identified by CCDR~\cite{aragam2015concave}, MMPC (Max-Min Parents-Children)~\cite{tsamardinos2003time}, GES (Greedy Equivalence Search)~\cite{chickering2002optimal}, and PC (Peter-Clark)~\cite{colombo2014order}. We use the implementations provided by the  pcalg R package~\cite{Hauser2012MarkovEquiv,Kalisch2012pcalg} and causal discovery toolbox (CDT)~\cite{Kalainathan2019Causal}\footnote{\url{https://fentechsolutions.github.io/CausalDiscoveryToolbox/html/index.html}}. The outcomes of our ensemble approach to causal discovery is a causal graph reflecting the relationships within the causal system, weighting edges by the agreement among the individual algorithms on whether the causal relationship exists. 

After applying this causal discovery approach to each year individually, we are able to construct a dynamic causal graph and investigate trends in causal relationships -- \eg as they are introduced, persist over time, or are eliminated.

\subsection{Treatment Effect Estimation}
We further investigate the magnitude and effect of causal relationships using average treatment effect (ATE) estimates. We compare pair-wise estimates using several causal inference models: {\it Causal Forest}~\cite{tibshirani2018package} and {\it Propensity Score Matching}~\cite{ho2007matching} using the ``MatchIt'' R package\footnote{\url{https://cran.r-project.org/web/packages/MatchIt/vignettes/MatchIt.html}}, and a cluster-based conditional treatment effect estimation tool -- {\it Visualization and Artificial Intelligence for Natural Experiments (VAINE)}\footnote{\url{https://github.com/pnnl/vaine-widget}}~\cite{guo2021vaine}. 

\begin{figure*}[t!]
    \centering
    \includegraphics[width=0.67\textwidth]{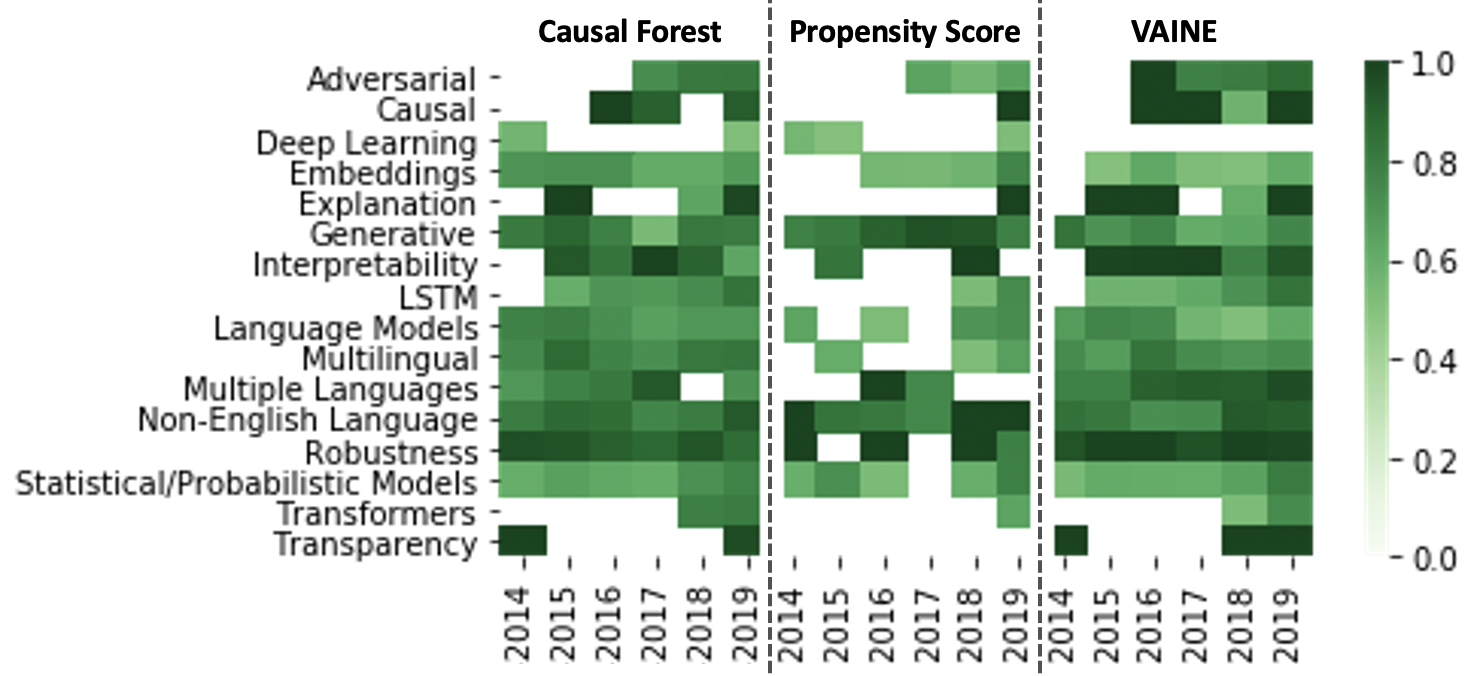}  
    \caption{Treatment effect estimates obtained using three causal inference methods -- Causal Forest, Propensity Score Matching and VAINE, for publish on $x$ $\rightarrow$ retire $x$ over time, across TEE methods.}
    \label{fig:publish_x_retire_x_tee}
\end{figure*}

\begin{figure*}[t!]
    \centering
    \includegraphics[width=0.49\textwidth]{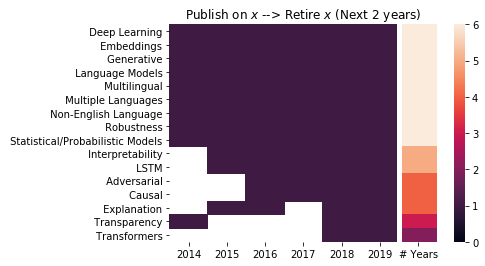}
    \includegraphics[width=0.49\textwidth]{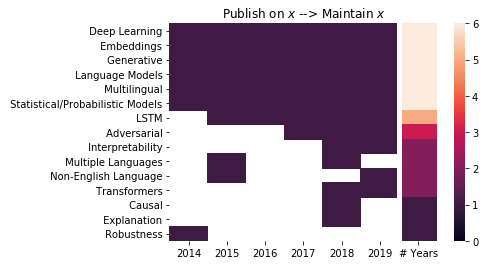} 
    \caption{Summary of causal structure learning using our ensemble model discovered from Publish on $x$ to Retire $x$ (above) or Maintain $x$ in the next 2 years (below), by year. Shaded cells indicate that an edge was discovered, white cells indicate that no edge was discovered for that year. At right is a summary of the number of years for which an edge was discovered.}
    \label{fig:publish_x_future_x}
    \vspace{-0.25cm}
\end{figure*}

VAINE is designed to discover natural experiments and estimate causal effects using observational data and address challenges traditional approaches have with continuous treatments and high-dimensional feature spaces. First, VAINE allows users to automatically detect sets of observations controlling for various covariates in the latent space. Then, using linear modeling, VAINE allows to estimate the treatment effect within each group and then average these local treatment effects to estimate the overall average effect between a given treatment and outcome variable. VAINE's novel approach for causal effect estimation allows it to handle continuous treatment variables without arbitrary discretization and produces results that are intuitive, interpretable, and verifiable by a human. VAINE is an interactive capability that allows the user to explore different parameter settings such as the number of groups, the alpha threshold to identify significant effects, etc.

Below we define what we mean by learning a causal effect from observational data. Given $n$ instances $[(x_1, t_1),\dots,(x_n, t_n)]$ learning causal effects quantifies how the outcome $y$ is expected to change if we modify the treatment from $c$ to $t$, which can be defined as $\mathbb{E}(y \mid t) - \mathbb{E}(y \mid c)$, where $t$ and $c$ denote a treatment as a control.

Similarly to our causal discovery based analyses, we examine the growth and decay of causal influence for a series of treatments (research focus represented by materials, methodology, or application-based keywords) on our outcomes of interest.

\subsection{Evaluation}
Evaluating causal analysis methods is challenging~\cite{saldanhaevaluation,weld2020adjusting,gentzel2019case,shimoni2018benchmarking,mooij2016distinguishing,dorie2019automated,singh2018comparative,raghu2018evaluation}. Broadly, evaluation techniques include structural, observational, interventional and qualitative techniques e.g., visual inspection. Observational evaluation by nature are non-causal and do not have the ability to measure the errors under interventions. Structural measures are limited due to the requirements of known structure, are oblivious to magnitude and type and dependence, as well as treatments and outcomes, and constrain research directions. Unlike structural and observational measures, interventional measures allow to evaluate model estimates of interventional effects e.g., ``what-if counterfactual evaluation". 

In this work we rely on both qualitative and quantitative evaluation. Methods that we use for causal inference were independently validated using structural and observational measures on synthetic datasets -- causal forest~\cite{wager2018estimation}, propensity score matching~\cite{causaleval}, causal ensemble~\cite{saldanhaevaluation}, and VAINE~\cite{guo2021vaine}. Since we rely on four complementary causal inference techniques, we draw our conclusions based on their agreement. In addition, to perform qualitative evaluation with the human in the loop we rely on recently release visual analytics tools to evaluate causal discovery and inference~\cite{causaleval}.

\section{Results}
In this section, we present a series of key findings surfaced within the causal mechanisms discovered and treatment effects estimated focusing on contrastive analysis over time: 
whether publishing on a given research entity (methodology, task, material) influences continuing to publish in that area or with that methodology, how authors shift from existing to novel methodology over time, and evidence of external events (\ie funding or large research programs) potentially impacts the adoption and maintenance of publication trends. 

\begin{figure*}[t!]
    \centering
    \includegraphics[width=0.65\textwidth]{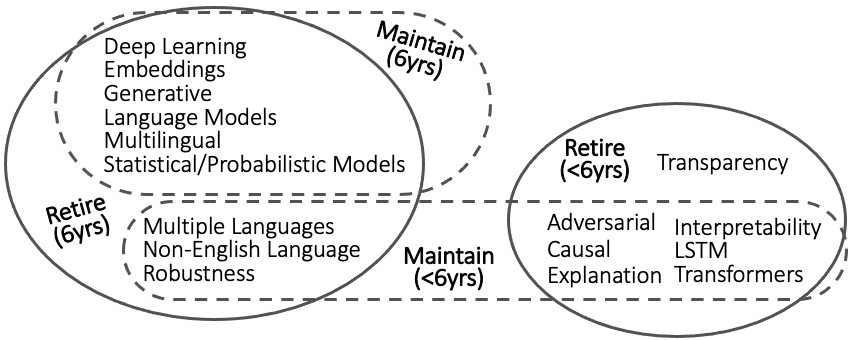} 
    \caption{Venn diagram illustrating shared research focus among causal relationships discovered by multiple causal inference methods in Figure~\ref{fig:publish_x_future_x}. This plot demonstrates long-lasting vs. short-lasting research trends in the CL community.}
    \label{fig:publish_x_future_x_venn}
\end{figure*}

\subsection{Continuing Existing Avenues of Research}

One of the first trends we noticed, in both the causal structures and treatment effects, was a causal relationship between publishing on a given research entity (\eg robustness, LSTMs, transformers, NER, etc.) and whether an author would \textit{continue} to publish on the same topic, task, or methodology in the following year(s). Does publishing once influence whether you will publish again? In short, no. We see a consistent trend in \textit{positive} treatment effects, as illustrated in Figure~\ref{fig:publish_x_retire_x_tee}, from publishing in the current year to not
publishing (pausing or retiring research entities) in the future -- publishing on $x$ leads to \textit{not publishing} on $x$ in the future.

In Figure~\ref{fig:publish_x_future_x}, we summarize the temporal dynamics of causal relationships from publishing on $x$ in a current year's publication to retiring $x$ (no publications associated with research entity $x$) in the next 2 years (above) or maintaining $x$ (at least one publication associated) in the next 2 years (below) indicated by our causal structural learning analyses which aligns with our TEE results. We show the consistency in which research entities are included in these trends using Figure~\ref{fig:publish_x_future_x_venn}. We see that many of the elements where causal relationships were identified in all 6 years are present in both the retirement and maintenance relationships. Of all the elements, research on \textit{Transparency} is the only case where there is only a retirement relationship. All elements with identified maintenance relationships in at least one year were also present in the set of retirement relationships.

\subsection{Emerging Research Foci, and the Impacts on Retirement of Old Research Foci}
 
\label{sec:churn}

The introduction or popularization of new model architectures (especially in deep learning) has an initial strong impact on retirement of previous SOTA architectures, but this is often focused on the initial adoption. We investigate several examples of such phenomena.  
Table~\ref{tab:lstm_bilstm} illustrates the decaying causal influence that using bidirectional LSTM-based architectures in current publications has on the retirement of (no longer using) LSTM in future publications. At first, there is a strong causal effect (approx. 0.8), where the use of biLSTM layers lead to no longer using LSTM layers. However, this reduces over time, with CF estimating close to no effect past 2015. We see a complementary trend on the relative publication volume increase outcome (\textit{Increase \# publications next year}), where there is an initial strong effect (0.76) that decays until it shifts to a negative effect (in 2018) then neutral (in 2019), as shown in Table~\ref{tab:lstm_bilstm}.

\begin{table}[t]
    \centering
    \small
    \setlength\tabcolsep{5 pt} 
    \begin{tabular}{l|ccccccc}
        \hline
          Method & 2014 & 2015 & 2016 & 2017 & 2018 & 2019 \\
           \hline
         CF & 0 & 0.71 \cellcolor{posColor!71} & -0.01 \cellcolor{negColor!1} & 0.07 \cellcolor{posColor!7} & 0.09  \cellcolor{posColor!9} &-0.03 \cellcolor{negColor!3}\\
         VAINE & 0 & 0.88 \cellcolor{posColor!88}& 0.46 \cellcolor{posColor!46} & 0.68 \cellcolor{posColor!68} & 0.68 \cellcolor{posColor!68} &0.77 \cellcolor{posColor!77}\\
         \hline
         \textit{Mean} & 0 & 0.80 \cellcolor{posColor!80}& 0.22 \cellcolor{posColor!22}& 0.32 \cellcolor{posColor!32}& 0.39 \cellcolor{posColor!39}&0.37 \cellcolor{posColor!37}\\
         \hline
         
    \end{tabular}
    \caption{Treatment effect estimates for the treatment \textit{Publish on bidirectional LSTM} on outcome \textit{Retire LSTM} by year, illustrating a decaying influence.}
    \label{tab:lstm_bilstm}
\end{table}

\begin{table}[t]
    \centering
    \small
    \setlength\tabcolsep{5 pt} 
    \begin{tabular}{l|ccccccc}
        \hline
          Method & 2014 & 2015 & 2016 & 2017 & 2018 & 2019 \\
           \hline 
           CF &
           0 \cellcolor{posColor!0} & 
           0.76 \cellcolor{posColor!76} & 
           -0.03 \cellcolor{negColor!3} & 
           0.36 \cellcolor{posColor!36} & 
           -0.2 \cellcolor{negColor!20} & 
           0.00
           \\
           VAINE &
           0 &
           0 &
           0 &
           0.39 \cellcolor{posColor!39} & 
           -0.23 \cellcolor{negColor!23} & 
           0 
           \\
           \hline
           \textit{Mean} &
           0 &
           0.38 \cellcolor{posColor!38} & 
           -0.02 \cellcolor{negColor!2} & 
           0.38  \cellcolor{posColor!38} & 
           -0.22  \cellcolor{negColor!22} & 
           0.00
           \\ 
         \hline
         
    \end{tabular}
    \caption{Treatment effect estimates for the treatment \textit{Publish on bidirectional LSTM} on outcome \textit{Increase Publications next year}, illustrating a strong initial influence shift to negative (2018) then neutral (2019).
    }
    \label{tab:lstm_bilstm}
    \vspace{-0.2cm}
\end{table}

\begin{figure*}[ht]
    \centering
    \includegraphics[width=\textwidth, trim={0 0 0 0.75cm},clip]{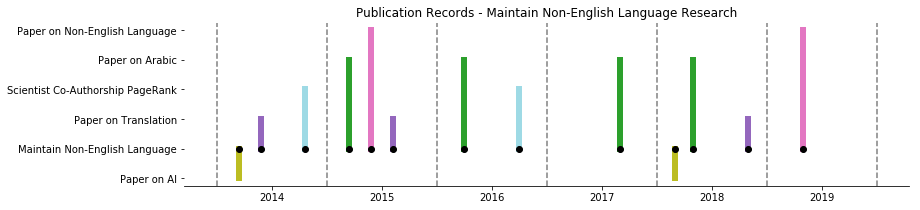}
    \caption{Recurrent causal relationships (identified for at least two years) that influence {\bf continued publication patterns} related to non-English languages in the CL community, \eg scientist co-authorship PageRank effects maintaining non-English publication focus in 2014 and 2016. Black markers identify the effect, with line segments extending to the cause nodes, and distinct relationships are represented by varying colors.} 
    \label{fig:cd_maintain_nonenglish}
        \vspace{-0.3cm}
\end{figure*}

\begin{figure*}[ht]
    \centering
    \includegraphics[width=0.8\textwidth, trim={0 0 0 0.75cm},clip]{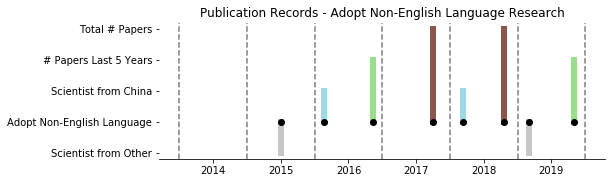}
    \caption{Recurrent causal relationships  
    that influence {\bf new publication patterns} related to non-English languages. 
    Black markers identify the effect, with line segmnets extending to the cause nodes, and distinct relationships are represented by varying colors. Empty time windows indicate no recurrent relationships were discovered.
    } 
    \label{fig:cd_adopt_nonenglish}
\end{figure*}

In addition, we see a consistent divergence from the trend described above (publishing on \textit{x} influences not publishing on \textit{x} in the next two years) for research related to non-English languages in the 2016-2018 time frame (see Figure~\ref{fig:cd_maintain_nonenglish}). These might be explained by the impact of large-scale research programs and funding (note, we will empirically confirm or dispute this hypothesis as a part of our future work). For example, we find that these outcomes (whether researchers continue to publish research related to non-English languages in 2017-2020) align with the last few years (and program-wide evaluation events\footnote{\url{https://www.nist.gov/itl/iad/mig/lorehlt-evaluations}}) of the LORELEI (Low Resource Languages for Emergent Incidents) DARPA program\footnote{\url{https://www.darpa.mil/program/low-resource-languages-for-emergent-incidents}}.

The goal of the LORELEI program was ``to dramatically advance the state of computational linguistics and human language technology to enable rapid, low-cost development of capabilities for low-resource languages'', and resulted in several publications on such languages from performers \eg \cite{strassel2016lorelei}. ``What is funded is published'' may be an intuitive influence, but here we see qualitative evidence that funding could influence the causal mechanisms of the publication ecosystem --- these signals are strong enough to be reflected in causal systems discovered using causal discovery algorithms in observational data. For \textit{adopting} non-English as a research focus, we also see influence from the authors' country associations -- \ie institution affiliations in China influence adopting non-English research (Fig.~\ref{fig:cd_adopt_nonenglish}). 

\begin{table}[t!]
    \centering
    \small
    \setlength\tabcolsep{5 pt} 
    Publish on Arabic $\rightarrow$ Continue Publishing on non-English
    \begin{tabular}{l|ccccccc} 
        \hline
          Method & 2014 & 2015 & 2016 & 2017 & 2018 & 2019 \\
           \hline
           
            CF & 0.04 \cellcolor{posColor!4} & 0.03 \cellcolor{posColor!3} & 0.28 \cellcolor{posColor!28} & 0.56 \cellcolor{posColor!56} & 0.40 \cellcolor{posColor!40} &-0.55\cellcolor{negColor!55}
            \\
            VAINE & 0 & 0.20 \cellcolor{posColor!20} & 0.43 \cellcolor{posColor!43} & 0.51 \cellcolor{posColor!51} & 0.13  \cellcolor{posColor!13} &0.06 \cellcolor{posColor!6}
            \\
            \hline
            \textit{Mean} & 0.02 \cellcolor{posColor!2} & 0.12 \cellcolor{posColor!12} & 0.36 \cellcolor{posColor!36} & 0.54 \cellcolor{posColor!54} & 0.27 \cellcolor{posColor!27} &0.00 \cellcolor{posColor!0} \\  \hline

        \multicolumn{6}{l}{}\\
            
    \end{tabular}
        \vspace{-0.25cm}
    
    Publish on Arabic $\rightarrow$ Stop Publishing on non-English
    \begin{tabular}{l|cccccc}
        \hline
          Method & 2014 & 2015 & 2016 & 2017 & 2018 & 2019 \\ \hline
           
            CF & 0.44 \cellcolor{posColor!44} & 0.80 \cellcolor{posColor!80} & 0.40 \cellcolor{posColor!40} & 0.13 \cellcolor{posColor!13} & 0.50 \cellcolor{posColor!50} &0.91 \cellcolor{posColor!91}
            \\
            VAINE & 0.81 \cellcolor{posColor!81} & 0.71 \cellcolor{posColor!71} & 0.45 \cellcolor{posColor!45} & 0.20 \cellcolor{posColor!20} & 0.82 \cellcolor{posColor!82} &0.91 \cellcolor{posColor!91}
            \\
            \textit{Mean} & 0.62 \cellcolor{posColor!62} & 0.75 \cellcolor{posColor!75} & 0.42 \cellcolor{posColor!42} & 0.16 \cellcolor{posColor!16} & 0.66 \cellcolor{posColor!66} &0.91 \cellcolor{posColor!91}
            \\

            \hline
        \multicolumn{6}{l}{}\\

    \end{tabular}
        \vspace{-0.25cm}

    Publish on Arabic $\rightarrow$ Increase Publications next year
    \begin{tabular}{l|cccccc}
        \hline
          Method & 2014 & 2015 & 2016 & 2017 & 2018 & 2019 \\  \hline
           
            CF ~~~~~& -0.03 \cellcolor{negColor!3} & -0.03 \cellcolor{negColor!3} & 0.31 \cellcolor{posColor!31} & 1.8 \cellcolor{posColor!100} & 0.23 \cellcolor{posColor!23} & -0.09 \cellcolor{negColor!9}
            \\ 

            \hline

    \end{tabular}
    \caption{Treatment effect estimates for the treatment \textit{Publish on Arabic} for outcomes \textit{Maintain non-English in the next two years} (above) and  \textit{Stop publishing on non-English in the next year} (below) by year, illustrating a peak in influence continuing to publish in 2017.}
    \label{tab:arabiic_maintain_nonenglish}
        \vspace{-0.3cm}
\end{table}

\begin{figure}[t]
    \centering
    \small
    \begin{tikzpicture}
 
    \begin{axis} [ybar, bar width=20pt,
    height=1.2in,
    width=3in, 
    xtick = {2014,2015,2016,2017,2018,2019},
    xticklabels = {2014,2015,2016,2017,2018,2019},
    xmin=2013.5, xmax=2019.5, 
    ymin=0,ymax=18, 
    ylabel=\% authors,
    ylabel near ticks,
    axis y line*=left,
    axis x line*=bottom,
    title=\small Persistence of Authorship in non-English Research, 
    title style={at={(0.5,0.8)}}
    ]
    \addplot[no marks, fill=blue!30!white, blue!30!white] coordinates { 
    (2014, 2.290076)
    (2015, 10.619469)
    (2016,  8.800000)
    (2017, 15.189873)
    (2018,  4.672897)
    (2019, 7.619048) 
    };
    \end{axis}

    \end{tikzpicture} 
    \vspace{-0.2cm}
    \caption{  
    The percentage of authors who published on non-English languages in a given year, who also published on non-English languages in the following year.}
    \label{fig:pct_nonenglish}
        \vspace{-0.3cm}
\end{figure}
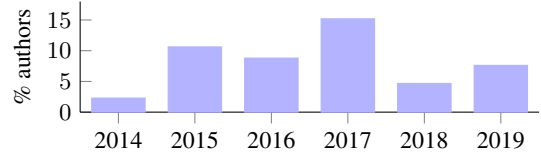

Table~\ref{tab:arabiic_maintain_nonenglish} illustrates a peak in the positive influence of publishing in a particular non-English language (\ie Arabic, which was one of the languages of interest for the LORELEI program) and continuing to publish on non-English languages. We see that the divergence of the causal relationships, and the persistence of authorship in non-English language research, illustrated in Figure~\ref{fig:pct_nonenglish}, center around or peak in 2017 and begin to flip (causal forest estimates a negative effect of -0.55) in 2019.

\section{Discussion and Conclusions}

In this study, we identified and analyzed causal influences between the trends and publishing behavior on future research performed and published in the computational linguistic community. Adjusting for confounders related to publication properties extracted from publication text and metadata, author characteristics, and collaboration network properties, we examine the causal relationships that could potentially be driving scientist productivity and the adoption, maintenance, and retirement of methods, materials, and tasks referenced in publications, and how these dynamics evolve over time.

Our analyses show that publishing once on a specific task, application, or methodology has a causal influence that causes authors not to publish on the same approach in the following year e.g., robustness, interpretability etc.  This is consistent across a significant number of the research methods, tasks, and application domains represented in our consolidated annotations. 

There are several potential drivers of this causal relationship. 
First, publishing in the CL community, particularly in recent years, is extremely competitive. Acceptance rates are low and the number of submissions each year continue to grow -- there is a lot of competition for few spots.
This could also be reflective of the churn in novelty and state-of-the-art technologies: as one technology (\eg LSTMs) are replaced by a new SOTA methodology (\eg biLSTMs, or transformers), this naturally leads to the retirement of the previous methods, which we also see reflected in the findings presented in \autoref{sec:churn}. As we drive the field forward, we stop publishing on older methods, materials, and tasks because of novelty incentives or requirements to be accepted in top-tier venues.

\paragraph{Limitations} A limitation of our current causal analysis approach is the restriction to ACL Anthology publication records only. As this dataset comprises only ACL venues, it does not guarantee inclusion of all possible publications from each author. For example, authors who publish in ACL venues may also publish in ICLR, NeurIPs, etc. Future work can address this limitation by augmenting the data set with supplementary collects from these venues (\eg using Google Scholar). 

Another avenue of future work is to incorporate funding information directly in the analyses. As we have shown, there is evidence that funding has causal relationships with publication outcomes and expertise evolution, and may act as a confounder for other relationships. Future work will extract funding from full text PDFs (\eg from the acknowledgements section) in order to adjust for the effects of funding as a confounder, which may also be an impactful treatment to analyze.

\section*{Acknowledgements}
This material is based on work funded by the United States Department of Energy (DOE) National Nuclear Security Administration (NNSA) Office of Defense Nuclear Nonproliferation Research and Development (DNN R\&D) Next-Generation AI research portfolio and Pacific Northwest National Laboratory, which is operated by Battelle Memorial Institute for the U.S. Department of Energy under contract DE-AC05-76RLO1830. Any opinions, findings, and conclusions or recommendations expressed in this material are those of the author(s) and do not necessarily reflect the views of the United States Government or any agency thereof. We would like to thank Joonseok Kim and Jasmine Eshun for their assistance preparing data.

\bibliographystyle{acl_natbib}
\bibliography{references,causal_eval_references,graph_comparison_references}

\newpage  
\appendix
\onecolumn
\section{Appendix}
\label{sec:appendix}
 
We summarize the publication record encodings (treatments, outcomes, and covariates)  used in our causal discovery and treatment effect estimation analyses in Table~\ref{tab:overview}.  

\begin{table}[h]
\small
    \centering
    \begin{tabular}{p{40mm}|p{110mm}}
         \hline
         Treatment & Description  \\
         \hline
         Publish on $x$ & Binary encoding of whether $x$ (one of our 50 research entities) was extracted from the publication's full text, using SpERT~\cite{luan2018multi}.  \\
         Scientist from $c$ & Binary encoding of whether author is a scientist affiliated with country $c$ (\ie author is associated with an insitution located in country $c$). There are six variations of this treatment, where $c$ is one of the five countries wiht the greatest representation in ACL publications (the United States, China, Germany, Japan, France) or ``Other''. \\  
         \hline 
    \end{tabular}
    \begin{tabular}{p{40mm}|p{110mm}}
        \multicolumn{2}{c}{}\\ 
         \hline
         Outcome & Description  \\
         \hline
         Adopt $x$ & Binary encoding of whether author will publish on $x$ for the first time in the next calendar year.\\
         Maintain $x$ & Binary encoding of whether author previously published on $x$ and has at least one publication on $x$ in the following calendar year.\\ 
         Retire/Pause $x$ & Binary encoding of whether author previously published on $x$ but has no publications on $x$ in the following two calendar years.\\
         Publication Increase Rate & The relative increase in the number of publications by author in the next calendar year, \ie $\frac{\#~publications~in~year~t+1}{\#~publications~in~year~t}$\\
         \hline 
    \end{tabular}
    \begin{tabular}{p{40mm}|p{110mm}}
        \multicolumn{2}{c}{}\\ 
         \hline 
         Covariate & Description  \\
         \hline
         \# Papers & Total number of author's publications (cumulative since first publication). \\
         \# Papers (Last Year) & Number of author's publications within the last year. \\
         \# Papers (Last 5 Years) & Number of author's publications within the last five years. \\
         \# Co-authors & Number of co-authors linked to author in the collaboration network (by year).\\
         Avg. \# Papers Co-authored & Average paper count per co-author, \ie $\frac{\# Papers}{\# Co-Authors}$ \\
         Co-author Likelihood & Likelihood that a randomly selected co-author was a co-author on the publication, \ie $\frac{Avg. \# Papers Co-authored}{\# Papers}$\\
         Centrality & Degree centrality of the author in the collaboration network, for the publication's calendar year. \\
         Page Rank & Page rank of the author in the collaboration network, for the publication's calendar year. \\
         Time Since First Paper & Number of years since the author's first publication in the ACL anthology. \\
         Conference & Binary encoding of whether the paper is published in a conference (1) or journal (0). \\
         \# Authors & Publication's total number of authors. \\
         Page Length & Page length of the publication's full text PDF. \\
         \# Words & Total number of words in the publication's full text.\\
         Gunning Fog Index & Gunning Fog Index readability measure~~\cite{gunning1952technique} calculated using the publication text. \\ 
         \hline
    \end{tabular}
    \caption{Overview of Treatments (above), Outcomes (center), and Covariates (below).}
    \label{tab:overview}
\end{table}

\end{document}